\documentclass[12pt]{article}
\usepackage[T1]{fontenc}
\fontfamily{georgia}\selectfont

\usepackage{lmodern, parskip, tikz, amsmath, cite, amssymb, mathtools, bbold, lettrine, bm, float, subcaption, wrapfig, mathtools, pgfplots}

\usetikzlibrary{arrows,chains,math}

\usepackage[labelformat=simple]{subcaption} 
\pgfplotsset{compat=1.15} 

\DeclarePairedDelimiter\autobracket{(}{)}
\newcommand{\br}[1]{\autobracket*{#1}}

\definecolor{burgundy}{rgb}{0.5, 0.0, 0.13}

\usepackage[colorlinks=true,
            linkcolor=burgundy,
            urlcolor=burgundy,
            citecolor=burgundy]{hyperref}

\colorlet{linkequation}{blue}

\newcommand*{\SavedEqref}{}
\let\SavedEqref\eqref
\renewcommand*{\eqref}[1]{%
  \begingroup
    \hypersetup{
      linkcolor=linkequation,
      linkbordercolor=linkequation,
    }%
    \SavedEqref{#1}%
  \endgroup
}
\usepackage[margin=1in]{geometry}
\title{
\vspace{-2em}
\LARGE \bf 
\rule{\textwidth}{3.5pt}
Elastic Weight Consolidation (EWC):\\ Nuts and Bolts\\ \rule{\textwidth}{1.4pt}
}
\author{Abhishek Aich \\[-0.3em]
\small{University of California, Riverside} \\[-0.4em]
\href{mailto: aaich@ece.ucr.edu}{\small{\texttt{aaich@ece.ucr.edu}}}}
\date{~}

\begin{document}
\maketitle
\vspace{-2em}
 
\begin{abstract}
\noindent\it In this report, we present a theoretical support of the continual learning method \textbf{Elastic Weight Consolidation}, introduced in paper titled `Overcoming catastrophic forgetting in neural networks' \cite{R1}. Being one of the most cited paper in regularized methods for continual learning, this report disentangles the underlying concept of the proposed objective function. We assume that the reader is aware of the basic terminologies of continual learning.
\end{abstract}

\section{Introduction}
Following are the notations used throughout this report. Vectors and matrices are denoted in bold lowercase and bold uppercase, respectively. Superscript $^\top$ denotes matrix transpose. $\mathbb{E}[\cdot]$ denotes the expectation operator. An optimum value of a variable is denoted by adding a superscript $^\star$. 

Continual learning is a much desired attribute for neural networks. For example, if we train a model to distinguish between images of a cat and a dog (task 1), and subsequently train it again to distinguish between images of chair and table (task 2), the model should be able to retain its knowledge on task 1 even after learning task 2. In simple terms, our network model should be able to perform equally well on all seen tasks, even after learning new ones. Any degradation of performance on the previous tasks after learning new ones is fittingly termed as \textit{catastrophic forgetting}. This sub-research area has seen an insurgence in works in recent times \cite{R1, zenke2017continual, li2017learning, aljundi2018memory}. Briefly, the continual learning scenarios can be categorized into following \cite{van2019three}:
\begin{itemize}
    \item \textbf{Task-Incremental Learning}: For the given set of tasks, the task identity is known during testing.
    
    \item \textbf{Domain-Incremental Learning}: For the given set of tasks, task identity is not provided during testing, but need not infer the same.
    
    \item \textbf{Class-Incremental Learning}: For the given set of tasks, task identity is not provided during testing, but has to infer the same.
\end{itemize}

We highly recommend \cite{van2019three, wiewel2019localizing} for a good overview of different methodologies to alleviate catastrophic forgetting as well as continual learning in general. The next Section describes the well studied regularization method of continual learning: Elastic Weight Consolidation. It presents a solution to the continual learning problem by making task-specific synaptic (\textit{read} network parameters) consolidation. Based on the theory of plasticity of post-synaptic dendritic spines in the brain, this method presents a paradigm that marks how important is a network parameter to the previous tasks and penalizes any change made to it depending upon the importance, while learning new tasks.
\newcommand{\boundellipse}[3]
{(#1) ellipse (#2 and #3)
}
\definecolor{darkpastelgreen}{rgb}{0.01, 0.75, 0.24}
\definecolor{cardinal}{rgb}{0.77, 0.12, 0.23}
\definecolor{calpolypomonagreen}{rgb}{0.12, 0.3, 0.17}
\definecolor{chocolate}{rgb}{0.48, 0.25, 0.0}
\tikzset{
  vlines/.style={
    path picture={
      \draw[xstep=#1, ystep=100cm, shift={(path picture bounding box.south west)} ]
      (path picture bounding box.south west) grid (path picture bounding box.north east);
    }
  }
}
\section{Elastic Weight Consolidation} 
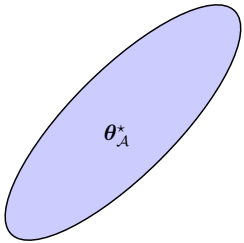
\begin{wrapfigure}{L}{0.3\textwidth}
    \centering
    \scalebox{0.7}{
    \begin{tikzpicture}
    
    \draw[thick, rotate = -45, fill=blue!20] \boundellipse{-2,4}{1}{3};

    \draw (1.3, 4.0) node {$\bm{\theta}^\star_{\mathcal{A}}$};  
    \end{tikzpicture}}
    \caption{
    \textbf{Possible configurations of} $\bm{\theta}^\star_{\mathcal{A}}$. The shaded region represents a space of optimum $\bm{\theta}_{\mathcal{A}}$ with acceptable errors w.r.t. $\bm{\theta}^\star_{\mathcal{A}}$ for task $\mathcal{A}$. 
    }
    \label{fig:fig1}
\end{wrapfigure}
Denote parameters of layers of a deep neural network (DNN) with $\bm{\theta}$. Training DNNs generates a mapping between the input distribution space and target distribution space. This is done by finding out an optimum $\bm{\theta} = \bm{\theta}^\star$ which results in the least error in the training objective. It has been shown in earlier works \cite{sussmann1992uniqueness} that such a mapping can be obtained with many configurations of $\bm{\theta}^\star$, represented in Fig.~\ref{fig:fig1}. The term \textit{many configurations} can be interpreted as a solution space around the most optimum $\bm{\theta}$ with acceptable error in the learned mapping. Note that in figures to follow, the shaded ellipses represent the solution of individual tasks where as the overlapping region of multiple ellipses, marked by \begin{tikzpicture}\draw[vlines = 1mm, thick] circle (0.2);\end{tikzpicture}, represents the common solution space for all tasks. 

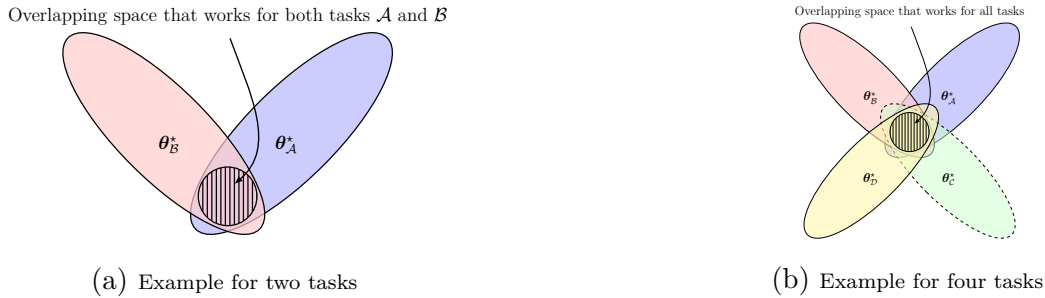
\begin{figure}[!htp]
    \begin{subfigure}[h]{0.45\textwidth}
    \centering
    \scalebox{0.6}{
    \begin{tikzpicture}
    \draw[thick, rotate = -45, fill=blue!20] \boundellipse{-2,4}{1}{3};
    
    \draw[thick, rotate = +45, fill=red!20, fill opacity=0.7] \boundellipse{+2,4}{1}{3};

    \draw (1.3, 4.0) node {$\bm{\theta}^\star_{\mathcal{A}}$};    
    \draw (-1.3, 4.0) node {$\bm{\theta}^\star_{\mathcal{B}}$};
    
    \node (a) at (0,6.5) {};
    \node (b) at (0,3) {};
    
    \draw[vlines = 1mm, thick] (0,2.85) circle (0.65);
    \draw[-latex, line width=0.3mm] (a) node[above] {Overlapping space that works for both tasks $\mathcal{A}$ and $\mathcal{B}$} to[out = -70, in = +40] (b) ;                 
    
    \end{tikzpicture}}
    \caption{\fontsize{8pt}{9pt}\selectfont
    Example for two tasks}
    \label{fig:fig2-1}
    \end{subfigure}
    \hfill
    \begin{subfigure}[h]{0.45\textwidth}
    \centering
    \scalebox{0.4}{
    \begin{tikzpicture}
    \draw[thick, rotate = -45, fill=blue!20] \boundellipse{-2,4}{1}{3};
    
    \draw[thick, rotate = +45, fill=red!20, fill opacity=0.7] \boundellipse{+2,4}{1}{3};
    
    \draw[thick, dashed, rotate = -135, fill=green!20, fill opacity=0.5] \boundellipse{-2,-0.2}{1}{3};
    
    \draw[thick, rotate = +135, fill=yellow!40, fill opacity=0.6] \boundellipse{2,-0.2}{1}{3};

    \draw (+1.3, 4.0) node {$\bm{\theta}^\star_{\mathcal{A}}$}; 
    \draw (-1.3, 4.0) node {$\bm{\theta}^\star_{\mathcal{B}}$};
    \draw (+1.3, 1.3) node {$\bm{\theta}^\star_{\mathcal{C}}$};
    \draw (-1.3, 1.3) node {$\bm{\theta}^\star_{\mathcal{D}}$};
    
    \node (a) at (0,6.5) {};
    \node (b) at (0,3) {};
    
    \draw[vlines = 1mm, thick] (0,2.85) circle (0.65);
    \draw[-latex, line width=0.3mm] (a) node[above] {Overlapping space that works for all tasks} to[out = -70, in = +40] (b) ;                            
    \end{tikzpicture}}
    \caption{\fontsize{8pt}{9pt}\selectfont
    Example for four tasks}
    \label{fig:fig2-2}
    \end{subfigure}
\caption{
\textbf{Overlap of possible configurations of} $\bm{\theta}^\star$. The overlapping space represents an optimum parameter region where the network performs without any catastrophic degradation on previous tasks.}
\end{figure}

Let's begin with a simple case of two tasks, task $\mathcal{A}$ and task $\mathcal{B}$. To have a configuration of parameters that performs well for both $\mathcal{A}$ and $\mathcal{B}$, the network should be able to pick $\bm{\theta}$ from the overlapping region of the individual solution spaces (see Fig.~\ref{fig:fig2-1}). This is with the assumption that there is always an overlapping region for the solution spaces of all tasks for the network to learn them sequentially. A case of four tasks has been illustrated in Fig.~\ref{fig:fig2-2}. In the first instance, the network can learn any $\bm{\theta} = \bm{\theta}_\mathcal{A}$ that performs well for task $\mathcal{A}$. But with the arrival of task $\mathcal{B}$, the network should pick up a $\bm{\theta} = \bm{\theta}_{\mathcal{A}, \mathcal{B}}$.
The next question that arrives is how can the network learn the a set of parameters that lies in this overlapping region. To this end, EWC presents a method of selective regularization of parameters $\bm{\theta}$. After learning $\mathcal{A}$, this regularization method identifies which parameters are important for $\mathcal{A}$, and then penalizes any change made to the network parameters according to their importance while learning $\mathcal{B}$. 

To formulate the objective, we start by taking a Bayesian approach needed to estimate the network parameters $\bm{\theta}$. More specifically given the data $\bm{\Sigma}$, we want to learn the posterior probability distribution function $p\br{\bm{\theta}\vert\bm{\Sigma}}$. Following \cite{RotateEWC} and using Bayes rule, write  
\begin{align}\label{eq:eq1}
    \underbrace{p\br{\bm{\theta}\vert\bm{\Sigma}}}_{\text{posterior}} = \dfrac{\overbrace{p\br{\bm{\Sigma}\vert\bm{\theta}}}^{\text{likelihood}}\overbrace{p\br{\bm{\theta}}}^{\text{prior}}}{p\br{\bm{\Sigma}}}
\end{align}
Since maximizing a function is same as maximizing its logarithm, we take $\log{\br{\cdot}}$ of \eqref{eq:eq1} as follows.
\begin{align}\label{eq:eq2}
    \log\br{p\br{\bm{\theta}\vert\bm{\Sigma}}} = \log\br{p\br{\bm{\Sigma}\vert\bm{\theta}}} +\log\br{p\br{\bm{\theta}}} - \log\br{p\br{\bm{\Sigma}}}
\end{align}
To train the neural network on $\bm{\Sigma}$, the objective function to be optimized over the log-likelihood function. 
\begin{align}\label{eq:eq3}
    \text{arg}\max_{\bm{\theta}}~\Big{\{}\ell\br{\bm{\theta}} = \log\br{p\br{\bm{\theta}\vert\bm{\Sigma}}}\Big{\}}
\end{align}
For the case of given two independent tasks such that $\bm{\Sigma} = \big{\{}\mathcal{A}, \mathcal{B}\big{\}}$ (with $\mathcal{B}$ appearing in sequence after $\mathcal{A}$), \eqref{eq:eq2} can be written as
\begin{align}\label{eq:eq4}
    \log\br{p\br{\bm{\theta}\vert\bm{\Sigma}}} 
    &= \log\br{p\br{\mathcal{B}\vert \mathcal{A}, \bm{\theta}}} +\log\br{p\br{\bm{\theta}\vert\mathcal{A}}} - \log\br{p\br{\mathcal{B}\vert \mathcal{A}}}\nonumber\\
    &= \log\br{p\br{\mathcal{B}\vert\bm{\theta}}} +\log\br{p\br{\bm{\theta}\vert\mathcal{A}}} - \log\br{p\br{\mathcal{B}}}\quad\br{\because\text{$\mathcal{A}$ and $\mathcal{B}$ are independent}}
\end{align}
Following \eqref{eq:eq1}, $p\br{\mathcal{B}\vert\bm{\theta}}$ is the loss for current task $\mathcal{B}$, $p\br{\mathcal{B}}$ is the likelihood for $\mathcal{B}$, and now posterior $p\br{\bm{\theta}\vert\mathcal{A}}$ for $\mathcal{A}$ becomes prior for $\mathcal{B}$.     
\pgfplotsset{compat=1.14}
\tikzmath{%
  function h1(\x, \lx) { return (9*\lx + 3*((\lx)^2) + ((\lx)^3)/3 + 9); };
  function h2(\x, \lx) { return (3*\lx - ((\lx)^3)/3 + 4); };
  function h3(\x, \lx) { return (9*\lx - 3*((\lx)^2) + ((\lx)^3)/3 + 7); };
  function skewnorm(\x, \l) {
    \x = (\l < 0) ? -\x : \x;
    \l = abs(\l);
    \e = exp(-(\x^2)/2);
    return (\l == 0) ? 1 / sqrt(2 * pi) * \e: (
      (\x < -3/\l) ? 0 : (
      (\x < -1/\l) ? \e / (8 * sqrt(2 * pi)) * h1(\x, \x*\l) : (
      (\x <  1/\l) ? \e / (4 * sqrt(2 * pi)) * h2(\x, \x*\l) : (
      (\x <  3/\l) ? \e / (8 * sqrt(2 * pi)) * h3(\x, \x*\l) : (
      sqrt(2/pi) * \e)))));
  };
}
\pgfmathdeclarefunction{gauss}{2}{%
  \pgfmathparse{1/(#2*sqrt(2*pi))*exp(-((x-#1)^2)/(2*#2^2))}%
}
\subsection{Intractability of posterior of $\mathcal{A}$ and it's approximation}
\begin{wrapfigure}{R}{0.4\textwidth}
\centering
\scalebox{0.7}{
\begin{tikzpicture}
\begin{axis}[every axis plot post/.append style={
  mark=none,domain=-4:3,samples=100,smooth}, 
  axis y line=none,
    axis x line*=bottom,
    clip=false,
    ymin=0,
    xtick=\empty,
    legend style={at={(1.1,0.9)},anchor=north east}] 
  \addplot[calpolypomonagreen, dashed, line width=1.5pt] {gauss(-0.4,0.5)};
  \addplot[chocolate, line width=1.5pt]    {skewnorm(x, -4)};
  \path (-0.4,0) coordinate (MM);
  \draw[line width=1.0pt, black] (-0.4,0) -- (-0.4,0.8);
  \legend{$\mathcal{N}\br{\bm{\theta}^\star_{\mathcal{A}}, [\mathbb{I}_{\mathcal{A}}]^{-1}}$, posterior pdf}
  \draw [yshift=-0.6cm, latex-latex](axis cs:-1.15,0.3) -- node [fill=white] {$[\mathbb{I}_{\mathcal{A}}]^{-1}$} (axis cs:0.35,0.3);
\end{axis}
\draw[latex-] (MM) --++ (0,-0.4) node[below,inner
sep=1pt]{$\bm{\theta}^\star_{\mathcal{A}}$};
\end{tikzpicture}}
\caption{
\textbf{Laplace approximation of true posterior pdf.} $\mathbb{I}_\mathcal{A}$ represents the Fisher Information matrix. See Section \ref{sec:sec22} for more. \vspace{-0.75em}}
\label{fig:fig3-1}
\end{wrapfigure}
Referring \eqref{eq:eq4}, it can be observed that we have to deal with the function $p\br{\bm{\theta}\vert\mathcal{A}}$. This is the posterior function for $\mathcal{A}$ which contains the information about the parameters that explain $\mathcal{A}$ using the given network. As discussed in \cite{R1}, this posterior function is said to be intractable. Basically, the intractability of $p\br{\bm{\theta}\vert\mathcal{A}}$ can be interpreted as the function not existing in some interpretable form. Hence, it is difficult to estimate its quantiles. See \cite{PosteriorDuke} for an example.  

Next as the posterior is difficult to analyze in its present form, we aim to approximate it using Laplace approximation. In simple terms, Laplace approximation methodology is employed to find a normal distribution approximation to a continuous
probability density distribution (see Fig.~\ref{fig:fig3-1}). Assuming $p\br{\bm{\theta}\vert\mathcal{A}}$ is smooth and majorly peaked around its point of maxima (i.e. $\bm{\theta}^\star_{\mathcal{A}}$), we can approximate it with a normal distribution with mean $\bm{\theta}^\star_{\mathcal{A}}$ and variance $[\mathbb{I}_{\mathcal{A}}]^{-1}$. This brings us to the question on how did we come to the conclusion on these particular values of mean and variance for the normal distribution. 

To begin with, compute the second order Taylor expansion of $\ell\br{\bm{\theta}}$ around $\bm{\theta}^\star_{\mathcal{A}}$ as follows. 
\begin{align}
    \ell\br{\bm{\theta}}\approx
    \ell\br{\bm{\theta}^\star_{\mathcal{A}}} +\br{ \dfrac{\partial\ell\br{\bm{\theta}}}{\partial\bm{\theta}}\bigg{\vert}_{\bm{\theta}^\star_{\mathcal{A}}}} + 
    \dfrac{1}{2}\br{\bm{\theta} -\bm{ \theta}^\star_{\mathcal{A}}}^\top\br{\dfrac{\partial^2\ell\br{\bm{\theta}}}{\partial^2\bm{\theta}}\bigg{\vert}_{\bm{\theta}^\star_{\mathcal{A}}}}\br{\bm{\theta} -\bm{ \theta}^\star_{\mathcal{A}}} \nonumber\\+ 
    \br{\text{higher order terms}}\cdots
\end{align}
Neglecting higher order terms and noting that $\dfrac{\partial\ell\br{\bm{\theta}}}{\partial\bm{\theta}}\bigg{\vert}_{\bm{\theta}^\star_{\mathcal{A}}} = 0$ (slope of tangent at peak), we have
\begin{align}\label{eq:eq6}
    \ell\br{\bm{\theta}}\approx
    \ell\br{\bm{\theta}^\star_{\mathcal{A}}} + 
    \dfrac{1}{2}\br{\bm{\theta} -\bm{ \theta}^\star_{\mathcal{A}}}^\top\underbrace{\br{\dfrac{\partial^2\ell\br{\bm{\theta}}}{\partial^2\bm{\theta}}\bigg{\vert}_{\bm{\theta}^\star_{\mathcal{A}}}}}_{\text{Hessian}}\br{\bm{\theta} -\bm{ \theta}^\star_{\mathcal{A}}}
\end{align}
Using \eqref{eq:eq3}, we can write \eqref{eq:eq6} for task $\mathcal{A}$ as following.
\begin{align}\label{eq:eq7}
    \log\br{p\br{\bm{\theta}\vert\mathcal{A}}} &= \log\br{p\br{\bm{\theta}^\star_{\mathcal{A}}\vert\mathcal{A}}} + \dfrac{1}{2}\br{\bm{\theta} -\bm{ \theta}^\star_{\mathcal{A}}}^\top\br{\dfrac{\partial^2\br{\log\br{p\br{\bm{\theta}\vert\mathcal{A}}}}}{\partial^2\bm{\theta}}\bigg{\vert}_{\bm{\theta}^\star_{\mathcal{A}}}}\br{\bm{\theta} -\bm{ \theta}^\star_{\mathcal{A}}}\nonumber\\
    \implies \log\br{p\br{\bm{\theta}\vert\mathcal{A}}} &= \dfrac{1}{2}\br{\bm{\theta} -\bm{ \theta}^\star_{\mathcal{A}}}^\top\br{\dfrac{\partial^2\br{\log\br{p\br{\bm{\theta}\vert\mathcal{A}}}}}{\partial^2\bm{\theta}}\bigg{\vert}_{\bm{\theta}^\star_{\mathcal{A}}}}\br{\bm{\theta} -\bm{ \theta}^\star_{\mathcal{A}}} + \Delta 
\end{align}
where $\Delta = \log\br{p\br{\bm{\theta}^\star_{\mathcal{A}}\vert\mathcal{A}}}$. \\Next, write $\br{\dfrac{\partial^2\br{\log\br{p\br{\bm{\theta}\vert\mathcal{A}}}}}{\partial^2\bm{\theta}}\bigg{\vert}_{\bm{\theta}^\star_{\mathcal{A}}}}$ as $-\br{\br{-\dfrac{\partial^2\br{\log\br{p\br{\bm{\theta}\vert\mathcal{A}}}}}{\partial^2\bm{\theta}}\bigg{\vert}_{\bm{\theta}^\star_{\mathcal{A}}}}^{-1}}^{-1}$ and replace it back in \eqref{eq:eq7} to express the same in the standard form of normal distribution function. \begin{align}\label{eq:eq8i}
    p\br{\bm{\theta}\vert\mathcal{A}} &= \epsilon\exp\br{-\dfrac{1}{2}\br{\bm{\theta} -\bm{ \theta}^\star_{\mathcal{A}}}^\top\br{\br{-\dfrac{\partial^2\br{\log\br{p\br{\bm{\theta}\vert\mathcal{A}}}}}{\partial^2\bm{\theta}}\bigg{\vert}_{\bm{\theta}^\star_{\mathcal{A}}}}^{-1}}^{-1}\br{\bm{\theta} -\bm{ \theta}^\star_{\mathcal{A}}}}
\end{align}
where $\epsilon = \exp\br{\Delta}$ is a constant. From \eqref{eq:eq8i}, it can be concluded that we have obtained the Laplace approximation of posterior pdf as
\begin{align*}
    p\br{\bm{\theta}\vert\mathcal{A}}\sim\mathcal{N}\br{\bm{ \theta}^\star_{\mathcal{A}}, \br{-\dfrac{\partial^2\br{\log\br{p\br{\bm{\theta}\vert\mathcal{A}}}}}{\partial^2\bm{\theta}}\bigg{\vert}_{\bm{\theta}^\star_{\mathcal{A}}}}^{-1}}
\end{align*} 
This has been illustrated in Fig.~\ref{fig:fig3-1}.
\subsection{Importance of parameters using Fisher Information matrix}\label{sec:sec22}
Notice the variance of the estimated normal distribution of $p\br{\bm{\theta}\vert\mathcal{A}}$. Given  $\bm{ \theta}^\star_{\mathcal{A}}$, the term $\log\br{p\br{\bm{\theta}\vert\mathcal{A}}}$ represents the log-likelihood of posterior pdf $p\br{\bm{\theta}\vert\mathcal{A}}$. Clearly, the term represents the inverse of \textbf{Fisher information matrix} (FIM), $\mathbb{I}_{\mathcal{A}} = \mathbb{E}\Bigg{[}-\dfrac{\partial^2\br{\log\br{p\br{\bm{\theta}\vert\mathcal{A}}}}}{\partial^2\bm{\theta}}\bigg{\vert}_{\bm{\theta}^\star_{\mathcal{A}}}\Bigg{]}$. Note that we obtain $\mathbb{I}_{\mathcal{A}}$ by using \eqref{eq:eq1} and treating the prior $p(\bm{\theta})$ and $p(\mathcal{A})$ constant. This makes derivative of log of \eqref{eq:eq1} posterior and likelihood equal. More on this in \textbf{Appendix} \textbf{A.2} of \cite{van2019three}. Finally, we get $p\br{\bm{\theta}\vert\mathcal{A}}\sim\mathcal{N}\br{\bm{ \theta}^\star_{\mathcal{A}}, [\mathbb{I}_{\mathcal{A}}]^{-1}}$. Further, as FIM can also be computed from first order derivatives, we can avoid the Hessian computed in \eqref{eq:eq6} using the following property \cite{kay1993fundamentals}.
\begin{align}\label{eq:eq9}
    \mathbb{I}_{\mathcal{A}} 
    &= \mathbb{E}\Bigg{[}-\dfrac{\partial^2\br{\log\br{p\br{\bm{\theta}\vert\mathcal{A}}}}}{\partial^2\bm{\theta}}\bigg{\vert}_{\bm{\theta}^\star_{\mathcal{A}}}\Bigg{]}\nonumber\\ 
    &= \mathbb{E}\Bigg{[}\br{\br{\dfrac{\partial\br{\log\br{p\br{\bm{\theta}\vert\mathcal{A}}}}}{\partial\bm{\theta}}}\br{\dfrac{\partial\br{\log\br{p\br{\bm{\theta}\vert\mathcal{A}}}}}{\partial\bm{\theta}}}^\top}\bigg{\vert}_{\bm{\theta}^\star_{\mathcal{A}}}\Bigg{]}
\end{align}
Putting \eqref{eq:eq7} back together in \eqref{eq:eq4}, we have
\begin{align}
    \log\br{p\br{\bm{\theta}\vert\bm{\Sigma}}} 
    &= \log\br{p\br{\mathcal{B}\vert\bm{\theta}}} +\log\br{p\br{\bm{\theta}\vert\mathcal{A}}} - \log\br{p\br{\mathcal{B}}}\nonumber\\
    &= \log\br{p\br{\mathcal{B}\vert\bm{\theta}}} +\dfrac{\lambda}{2}\br{\bm{\theta} -\bm{ \theta}^\star_{\mathcal{A}}}^\top\br{\dfrac{\partial^2\br{\log\br{p\br{\bm{\theta}\vert\mathcal{A}}}}}{\partial^2\bm{\theta}}\bigg{\vert}_{\bm{\theta}^\star_{\mathcal{A}}}}\br{\bm{\theta} -\bm{ \theta}^\star_{\mathcal{A}}} + \epsilon'
\end{align}
where $\epsilon'$ accounts for all constants and $\lambda$ is a hyper-parameter introduced to have a trade off between learning $\mathcal{B}$ and not forgetting $\mathcal{A}$. Simplifying more, we have
\begin{align}\label{eq:eq11}
    \log\br{p\br{\bm{\theta}\vert\bm{\Sigma}}}
    &= \log\br{p\br{\mathcal{B}\vert\bm{\theta}}} +
    \dfrac{\lambda}{2}\br{\bm{\theta} -\bm{ \theta}^\star_{\mathcal{A}}}^\top\br{\dfrac{\partial^2\br{\log\br{p\br{\bm{\theta}\vert\mathcal{A}}}}}{\partial^2\bm{\theta}}\bigg{\vert}_{\bm{\theta}^\star_{\mathcal{A}}}}\br{\bm{\theta} -\bm{ \theta}^\star_{\mathcal{A}}} + \epsilon'\nonumber\\
    &= \log\br{p\br{\mathcal{B}\vert\bm{\theta}}} -
    \dfrac{\lambda}{2}\br{\bm{\theta} -\bm{ \theta}^\star_{\mathcal{A}}}^\top\mathbb{I}_{\mathcal{A}}\br{\bm{\theta} -\bm{ \theta}^\star_{\mathcal{A}}} + \epsilon'\qquad\br{\text{Using \eqref{eq:eq9}}}\nonumber\\
    \implies \underbrace{\ell\br{\bm{\theta}}}_{\text{overall loss}} &= \underbrace{\ell_\mathcal{B}\br{\bm{\theta}}}_{\text{loss for $\mathcal{B}$ }} -
    \underbrace{\dfrac{\lambda}{2}\br{\bm{\theta} -\bm{ \theta}^\star_{\mathcal{A}}}^\top\mathbb{I}_{\mathcal{A}}\br{\bm{\theta} -\bm{ \theta}^\star_{\mathcal{A}}}}_{\text{weight regularizer}} + \epsilon'
\end{align}

Further simplification of \eqref{eq:eq11} can be found in \cite{van2019three}. Before we end this Section, let's discuss how does the FIM indicates the \textbf{importance} of the parameters for the previous tasks. 

We say a network has learnt a task when its objective has reached a minimum in the loss surface. We know that the curvature of such surfaces represent the sensitivity of the network with respect to the optimum $\bm{\theta}^\star$. This sensitivity can be determined by looking at the direction along which $\bm{\theta}^\star$ changes. This implies the curvature is inversely proportional to change in $\bm{\theta}^\star$. Hence, if the more the curvature, a `$\delta$' increment can result in large increase in the loss. Curvature of a curve is denoted by its Hessian and hence in our case, as the second derivative is of the log likelihood function of the posterior pdf, the FIM $\mathbb{I}_\mathcal{A}$ comes into picture. Thus, $\mathbb{I}_\mathcal{A}$ can tell us which parameter is important to the the previous task as its corresponding element in $\mathbb{I}_\mathcal{A}$ will have a large value, indicating higher importance. See \cite{maltoni2019continuous} for more.
\section{Conclusion}
The EWC methodology alleviates catastrophic forgetting by regularizing parameters of a network trained on previous tasks by penalizing any change in them according to their importance. This importance is indicated by the Fisher information matrix i.e. after a network is trained on one task, fine-tuning on the next task is performed on according to \eqref{eq:eq11}.
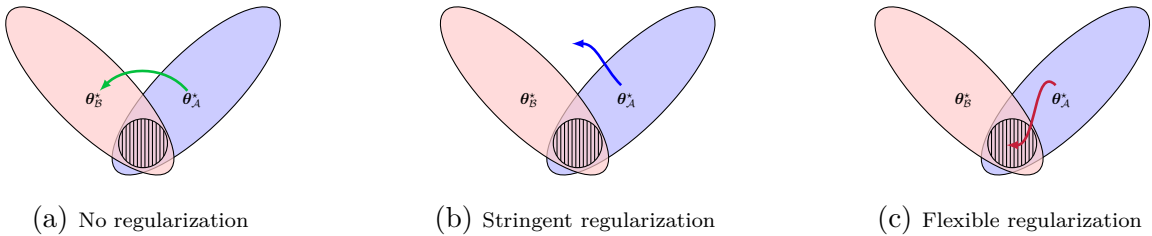
\begin{figure}[ht]
    \begin{subfigure}[h]{0.3\textwidth}
    \centering
    \scalebox{0.5}{
    \begin{tikzpicture}
    \draw[thick, rotate = -45, fill=blue!20] \boundellipse{-2,4}{1}{3};
    
    \draw[thick, rotate = +45, fill=red!20, fill opacity=0.7] \boundellipse{+2,4}{1}{3};

    \draw (1.3, 4.0) node {$\bm{\theta}^\star_{\mathcal{A}}$};    
    \draw (-1.3, 4.0) node {$\bm{\theta}^\star_{\mathcal{B}}$};
    
    \node (a) at (1.3, 4.1) {};
    \node (b) at (-1.3, 4.1) {};
    \draw[vlines = 1mm, thick] (0,2.85) circle (0.65);
    \draw[-latex, line width=0.7mm, color = darkpastelgreen] (a) node[below] {} to[out = +130, in = +45] (b) ;                            
    \end{tikzpicture}}
    \caption{\fontsize{8pt}{9pt}\selectfont
    No regularization}
    \label{fig:fig4-1}
    \end{subfigure}
    \hfill
    \begin{subfigure}[h]{0.3\textwidth}
    \centering
    \scalebox{0.5}{
    \begin{tikzpicture}
    \draw[thick, rotate = -45, fill=blue!20] \boundellipse{-2,4}{1}{3};
    
    \draw[thick, rotate = +45, fill=red!20, fill opacity=0.7] \boundellipse{+2,4}{1}{3};

    \draw (1.3, 4.0) node {$\bm{\theta}^\star_{\mathcal{A}}$};    
    \draw (-1.3, 4.0) node {$\bm{\theta}^\star_{\mathcal{B}}$};
    
    \node (a) at (1.3, 4.25) {};
    \node (b) at (-0.3, 5.5) {};
    \draw[vlines = 1mm, thick] (0,2.85) circle (0.65);
    \draw[-latex, line width=0.7mm, color = blue] (a) node[below] {} to[out = +135, in = 0] (b) ;                            
    \end{tikzpicture}}
    \caption{\fontsize{8pt}{9pt}\selectfont
    Stringent regularization}
    \label{fig:fig4-2}
    \end{subfigure}
    \hfill
    \begin{subfigure}[h]{0.3\textwidth}
    \centering
    \scalebox{0.5}{
    \begin{tikzpicture}
    \draw[thick, rotate = -45, fill=blue!20] \boundellipse{-2,4}{1}{3};
    
    \draw[thick, rotate = +45, fill=red!20, fill opacity=0.7] \boundellipse{+2,4}{1}{3};

    \draw (1.3, 4.0) node {$\bm{\theta}^\star_{\mathcal{A}}$};    
    \draw (-1.3, 4.0) node {$\bm{\theta}^\star_{\mathcal{B}}$};
    
    \node (a) at (1.3, 4.25) {};
    \node (b) at (-0.3, 2.8) {};
    
    \draw[vlines = 1mm, thick] (0,2.85) circle (0.65);
    \draw[-latex, line width=0.7mm, color = cardinal] (a) node[below] {} to[out = +135, in = 0] (b) ;                            
    \end{tikzpicture}}
    \caption{\fontsize{8pt}{9pt}\selectfont
    Flexible regularization}
    \label{fig:fig4-3}
    \end{subfigure}
\caption{
\textbf{Sequential training on task} $\mathcal{B}$ \textbf{after task} $\mathcal{A}$. \ref{fig:fig4-1}: Train the network as it is: results in `Forgetting', \ref{fig:fig4-2}: Make no change in the parameters of previous tasks, \ref{fig:fig4-3}: Make changes in the parameters of the previous tasks depending on their importance}
\label{fig:fig4}
\end{figure}
In the end, we refer to Fig.~\ref{fig:fig4} for a pictorial representation of EWC. Fig.~\ref{fig:fig4-1} represents the case when we simply fine-tune the network on the subsequent tasks. This makes the network learn the optimum parameters according to the current task, resulting in forgetting. Fig.~\ref{fig:fig4-2} represents the case when we apply a stringent regularization to the parameters learnt from the previous tasks. This method doesn't compute the importance of the parameters and penalizes all of them equally. This may result in the network not only forgetting the previous task, but also not be able to learn the current one. Finally, Fig.~\ref{fig:fig4-2} refers to the EWC methodology of computing the importance of parameters before fine-tuning on new tasks. This ensures that the network learns the optimum parameters that performs well for all tasks and hence, lies in the overlapping region of the solution spaces of the tasks in the given sequence.

\textbf{Acknowledgement} Thank you Tan Yan Rui
(e0441771@u.nus.edu) for suggestions.

\renewcommand{\refname}{\small{REFERENCES}}
\bibliographystyle{acm}
{\footnotesize
\bibliography{bibliography.bib}}
\end{document}